\documentclass[letterpaper,11pt]{article}
\usepackage{acl-hlt2011}
\usepackage{times}
\usepackage{latexsym}
\usepackage{graphicx}
\usepackage{url}
\usepackage{xspace}
\usepackage{relsize}
\newcommand{\secref}[2][]{Section#1~\ref{#2}\xspace}
\newcommand{\figref}[2][]{Figure#1~\ref{#2}\xspace}
\newcommand{\tabref}[2][]{Table#1~\ref{#2}\xspace}

\newcommand{\eqref}[1]{(\ref{#1})\xspace}

\newcommand{\mymeaning}[1]{``#1''}

\title{Large-Scale Noun Compound Interpretation \\
       Using Bootstrapping and the Web as a Corpus}

\author{Su Nam Kim \\
  Computer Science \& Software Engineering \\
  University of Melbourne\\
  Melbourne, VIC 3010\\
  Australia\\
  {\tt snkim@csse.unimelb.edu.au} \\\And
  Preslav Nakov\\
  Department of Computer Science\\
  National University of Singapore\\
  13 Computing Drive\\
  Singapore 117417\\
  {\tt nakov@comp.nus.edu.sg} \\}

\date{}

\begin{document}
\maketitle

\begin{abstract}

Responding  to the  need  for semantic  lexical  resources in  natural
language processing  applications, we examine methods  to acquire noun
compounds  (NCs), e.g.,  \emph{orange juice},  together with  suitable
fine-grained  semantic  interpretations, e.g.,  \emph{squeezed  from},
which  are directly  usable  as paraphrases.  We employ  bootstrapping
and  web statistics,  and  utilize the  relationship  between NCs  and
paraphrasing  patterns to  jointly extract  NCs and  such patterns  in
multiple alternating  iterations. In evaluation, we  found that having
one compound  noun fixed yields  both a higher number  of semantically
interpreted  NCs  and  improved  accuracy  due  to  stronger  semantic
restrictions.

\end{abstract}

\section{Introduction}
\label{sec:introduction}

Noun compounds (NCs) such as \emph{malaria mosquito} and \textit{colon
cancer tumor  suppressor protein} are challenging  for text processing
since  the relationship  between the  nouns  they are  composed of  is
implicit.
NCs are abundant in English and understanding their semantics is
important in many natural language processing (NLP) applications.
For example,  a question answering  system might need to  know whether
\emph{protein acting as  a tumor suppressor} is a  good paraphrase for
\emph{tumor  suppressor  protein}.  Similarly, a  machine  translation
system  facing the  unknown noun  compound \emph{Geneva  headquarters}
might  translate  it  better  if  it  could  first  paraphrase  it  as
\emph{Geneva headquarters of the WTO}.
Given  a  query for  \emph{``migraine  treatment''},  an
information  retrieval  system  could   use  paraphrasing  verbs  like
\emph{relieve}  and  \emph{prevent}  for query  expansion  and  result
ranking.

Most work on noun compound interpretation has focused on two-word NCs.
There have been two general lines of research:
the first one derives the NC semantics from the semantics of the nouns it is made of \cite{Rosario:2002,Moldovan:2004,Kim:2005,Girju:2007a,Oseaghdha:2009,Tratz:2010},
while the second one models the relationship between the nouns directly
\cite{Vanderwende:1994,Lapata:2002,Kim:2006,Nakov:2006:AIMSA,Nakov:2008,Butnariu:2008}.

In either  case, the semantics of  an NC is typically  expressed by an
abstract relation like \textsc{Cause} (e.g., \emph{malaria mosquito}),
\textsc{Source} (e.g.,  \emph{olive oil}), or  \textsc{Purpose} (e.g.,
\emph{migraine  drug}),  coming from  a  small  fixed inventory.  Some
researchers
however, have argued for a more fine-grained, even infinite, inventory \cite{Finin:1980}.
Verbs are particularly useful in this respect and can capture elements
of the semantics that the abstract relations cannot.
For  example,   while  most  NCs  expressing   \textsc{Make},  can  be
paraphrased by common patterns like \textit{be made of} and \textit{be
composed of},  some NCs allow  more specific patterns,  e.g., \emph{be
squeezed from} for \emph{orange juice},  and \emph{be topped with} for
\textit{bacon pizza}.

Recently,    the    idea    of   using    fine-grained    paraphrasing
verbs    for    NC    semantics   has    been    gaining    popularity
\cite{Butnariu:2008,Nakov:2008:AIMSA}; there  has also been  a related
shared  task   at  SemEval-2010   \cite{SemEval:2010Paraphrase}.  This
interest is partly  driven by practicality: verbs  are directly usable
as paraphrases.  Still, abstract relations remain  dominant since they
offer  a more  natural generalization,  which is  useful for  many NLP
applications.

One good  contribution to this debate  would be a direct  study of the
relationship between fine-grained and  coarse-grained relations for NC
interpretation. Unfortunately, the existing  datasets do not allow this
since they are tied to one particular granularity; moreover, they only
contain  a  few  hundred  NCs.  Thus, our  objective  is  to  build  a
large-scale dataset of hundreds of thousands of NCs, each interpreted
(1) by an abstract semantic relation and
(2) by a set of paraphrasing verbs.
Having such a large dataset would also help the overall advancement of the field.

Since there is no universally  accepted abstract relation inventory in
NLP,  and  since  we  are  interested in  NC  semantics  from  both  a
theoretical and  a practical viewpoint,  we chose the set  of abstract
relations  proposed in  the  theory of  \newcite{Levi:1978}, which  is
dominant  in theoretical  linguistics and  has been  also used  in NLP
\cite{Nakov:2008}.

We use a two-step algorithm to jointly harvest NCs and patterns (verbs
and prepositions) that  interpret them for a  given abstract relation.
First, we  extract NCs using  a small number  of seed patterns  from a
given abstract  relation. Then,  using the  extracted NCs,  we harvest
more patterns. This  is repeated until no new NCs  and patterns can be
extracted or for a pre-specified number of iterations.
Our  approach  combines  pattern-based extraction  and  bootstrapping,
which is novel for NC  interpretation; however, such combinations have
been used in other areas, e.g., named entity recognition
\cite{Riloff:1999,Thelen:2002,Curran:2007,McIntosh:2009}.

The remainder of the paper is organized as follows:
\secref{sec:relatedwork} gives an overview of related work,
\secref{sec:representation} motivates our semantic representation,
Sections \ref{sec:method}, \ref{sec:data}, and \ref{sec:experiment}
explain our method, dataset and experiments, respectively,
\secref{sec:discussion} discusses the results,
\secref{sec:erroranalysis} provides error analysis,
and \secref{sec:conclusion} concludes with suggestions for future work.

\section{Related Work}
\label{sec:relatedwork}

As we  mentioned above,  the implicit relation  between the  two nouns
forming a  noun compound can  often be expressed overtly  using verbal
and prepositional  paraphrases. For example, \textit{student  loan} is
\mymeaning{loan \underline{given to} a student}, while \textit{morning
tea} can be paraphrased as \mymeaning{tea \underline{in} the morning}.

Thus, many NLP approaches to NC semantics have used verbs and prepositions
as a fine-grained semantic representation
or as features when predicting coarse-grained abstract relations.
For  example,  \newcite{Vanderwende:1994} associated  verbs  extracted
from definitions in an online dictionary with abstract relations.
\newcite{Lauer:1995} expressed NC semantics using eight prepositions.
\newcite{Kim:2006} predicted abstract relations using verbs as features.
\newcite{Nakov:2008} proposed a fine-grained NC interpretation
using a distribution over Web-derived verbs, prepositions and coordinating conjunctions;
they also used this distribution to predict coarse-grained abstract relations.
\newcite{Butnariu:2008} adopted a similar fine-grained verb-centered approach to NC semantics.
Using a distribution over verbs as a semantic interpretation was also carried out in  a recent challenge: SemEval-2010  Task 9
\cite{butnariu-EtAl:2009:SEW,SemEval:2010Paraphrase}.

In noun compound interpretation, verbs and prepositions can be seen as patterns connecting the two nouns in a paraphrase. Similar pattern-based approaches have been popular in  information extraction  and  ontology   learning.
For example,
\newcite{Hearst:1992}   extracted hyponyms   using  patterns such as
\textit{X, Y,  and/or other  Zs},
where  \textit{Z} is a hypernym of \textit{X} and \textit{Y}.
\newcite{Berland:1999} used similar patterns to extract meronymy (part-whole) relations,
e.g., \textit{parts/NNS of/IN wholes/NNS} matches \textit{basements of buildings}.
Unfortunately, matches  are rare,  which makes  it difficult  to build
large  semantic inventories.  In  order to  overcome data  sparseness,
pattern-based approaches are often combined with bootstrapping.
For example, \newcite{Riloff:1999} used a multi-level bootstrapping
algorithm to learn both a semantic lexicon and extraction patterns,
e.g., \textit{owned by X} extracts \textsc{Company}
and \textit{facilities in X} extracts \textsc{Location}.
That is, they learned semantic lexicons using extraction patterns, and
then, alternatively, they extracted new patterns using these lexicons.
They also introduced a second level of bootstrapping to retain
the  most  reliable  examples  only.  While  the  method  enables  the
extraction of large lexicons,
its quality  degrades rapidly,  which makes it impossible  to run
for  too many  iterations.  Recently, \newcite{Curran:2007}  and
\newcite{McIntosh:2009}  proposed ways  to  control degradation  using
simultaneous learning and weighting.

Bootstrapping has  been applied to  noun compound extraction  as well.
For example, \newcite{Kim:2007}  used it to produce a  large number of
semantically interpreted noun compounds from  a small number of seeds.
In each iteration, the method replaced one component of an NC with its
synonyms, hypernyms and  hyponyms to generate a new NC.  These new NCs
were  further filtered  based on  their semantic  similarity with  the
original  NC.  While  the  method  acquired a  large  number  of  noun
compounds without significant semantic drifting, its accuracy degraded
rapidly  after each  iteration.
More importantly, the variation of the sense pairs was limited
since new NCs had to be semantically similar to the original NCs.

Recently,  \newcite{Kozareva:2010} combined patterns and bootstrapping
to learn the selectional restrictions for various semantic relations.
They used patterns involving the coordinating conjunction \textit{and},
e.g., ``\textit{*  and John  fly to *}'',  and learned  arguments such
as  \textit{Mary/Tom}  and  \textit{France/New  York}.  Unlike  in  NC
interpretation, it  is not  necessary for their  arguments to  form an
NC,  e.g.,  \textit{Mary  France}  and \textit{France  Mary}  are  not
NCs. Rather,  they were interested in building a semantic  ontology with  a predefined
set  of semantic  relations, similar to  YAGO  \cite{Suchanek:2007}, where
the  pattern \textit{work  for}  would have  arguments like  \textit{a
company/UNICEF}.

\section{Semantic Representation}
\label{sec:representation}

Inspired by \cite{Finin:1980},
\newcite{Nakov:2006:AIMSA} and \cite{Nakov:2008:AIMSA}
proposed that NC semantics is  best expressible
using paraphrases  involving verbs  and/or prepositions.  For example,
\textit{bronze  statue}  is  a  statue that  \textit{is  made  of,  is
composed of, consists of, contains, is of, is, is handcrafted from, is
dipped in, looks  like} bronze. They further proposed that selecting one
such  paraphrase is  not enough
and that multiple paraphrases are needed for a fine-grained  representation.
Finally, they observed that not all paraphrases are equally good
(e.g.,  \textit{is made  of}  is  arguably  better
than  \textit{looks like}  or \textit{is  dipped in}  for \textsc{Make}),
and thus proposed that the  semantics of  a noun  compound should be expressed
as  a  \emph{distribution} over  multiple possible paraphrases.
This line  of  research  was later adopted  by  SemEval-2010  Task  9
\cite{SemEval:2010Paraphrase}.

It easily  follows that the  semantics of abstract  relations
such as \textsc{Make} that can hold between  the nouns in an NC can be represented
in  the  same way:  as  a  distribution  over paraphrasing  verbs  and
prepositions.  Note, however, that some  NCs are paraphrasable  by
more  specific  verbs  that do not necessarily support the target abstract relation.
For  example, \textit{malaria  mosquito}, which  expresses
\textsc{Cause}, can  be paraphrased  using verbs  like \textit{carry},
which do not  imply direct causation. Thus, while we  will be focusing
on  extracting  NCs  for a  particular  abstract  relation,  we  are
interested in building semantic  representations that are specific for
these NCs  and do not  necessarily apply to \textit{all}  instances of
that relation.

Traditionally, the semantics  of a noun compound  have been represented
as  an abstract  relation  drawn  from a  small  closed  set.
Unfortunately, no such set is universally accepted, and mapping between sets has  proven challenging
\cite{Girju:2005}. Moreover, being both abstract and limited, such sets
capture  only  part of  the  semantics;  often multiple  meanings  are
possible, and sometimes none of  the pre-defined ones suits
a  given example. Finally, it  is unclear how useful  these sets are
since  researchers have often fallen  short of demonstrating practical uses.

Arguably, verbs have more expressive power and are more suitable for semantic representation:
there is an infinite number of them \cite{downing:1977:nc:sem},
and they can capture fine-grained aspects of the meaning. For example,
while both {\it wrinkle  treatment} and  {\it migraine  treatment} express
the  same abstract  relation \textsc{Treatment-For-Disease},
fine-grained differences can be revealed using verbs,
e.g., {\it  smooth} can paraphrase the former, but not the latter.

In many theories, verbs play an  important role in NC derivation  \cite{Levi:1978}.
Moreover, speakers often use verbs to make the hidden relation between the noun in a noun compound overt.
This allows for simple  extraction and for straightforward use
in NLP tasks like textual entailment \cite{Tatu:Moldovan:2005:entailment}
and machine translation \cite{Nakov:2008:ECAI}.

Finally, a single verb is often not enough,
and the meaning is better approximated by  a collection of verbs. For example,
while {\it malaria  mosquito} expresses \textsc{Cause}
(and is paraphrasable using {\it cause}),  further aspects of the meaning can  be captured
with more verbs, e.g., {\it carry}, {\it spread}, {\it be responsible for},
{\it be infected  with}, {\it transmit},  {\it pass on}, etc.

\section{Method}
\label{sec:method}

\begin{figure}[tbh]
\begin{center}
\includegraphics[width=2.8in]{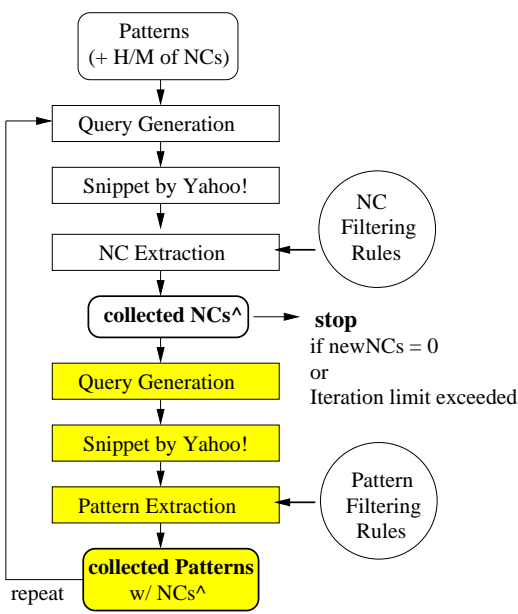}
\end{center}
\caption{Our bootstrapping algorithm.}
\label{fig:process}
\end{figure}

We   harvest   noun   compounds  expressing   some   target   abstract
semantic  relation   (in  the  experiments  below,   this  is  Levi's
\textsc{Make$_2$}),  starting  from a  small  number  of initial  seed
patterns: paraphrasing  verbs and/or prepositions. Optionally,  we might
also  be given  a  small  number of  noun  compounds that  instantiate
the  target   abstract relation.  We  then   learn  more  noun   compounds  and
patterns  for  the  relation  by  alternating  between  the  following
two  bootstrapping  steps,  using  the  Web as  a  corpus.  First,  we
extract more noun compounds that  are paraphrasable with the available
patterns  (see  \secref{sec:extraction-nc}).  We  then  look  for  new
patterns that  can paraphrase the newly-extracted  noun compounds (see
\secref{sec:extraction-pattern}). These  two steps are  repeated until
no  new noun  compounds can  be  extracted or  until a  pre-determined
number of iterations has been  reached. A schematic description of the
algorithm is shown in \figref{fig:process}.

\subsection{Bootstrapping Step 1: Noun Compound Extraction}
\label{sec:extraction-nc}

Given a list of patterns (verbs and/or prepositions),
we mine the Web to extract noun compounds that match these patterns.
We experiment with the following three bootstrapping strategies for this step:

\begin{itemize}
\item \textbf{Loose bootstrapping} uses the available patterns and imposes no further restrictions.
\item \textbf{Strict bootstrapping} requires that, in addition to the patterns themselves, some noun compounds matching each pattern be made available as well. A pattern is only instantiated in the context of either the head or the modifier of a noun compound that is known to match it.
\item \textbf{NC-only strict bootstrapping} is a stricter version of \emph{strict bootstrapping}, where the list of patterns is limited to the initial seeds.
\end{itemize}

\noindent Below we describe each of the sub-steps of the NC extraction process:
query generation, snippet harvesting, and noun compound acquisition \& filtering.

\subsubsection{Query Generation}

We generate generalized exact-phrase queries
to be used in a Web search engine (we use \emph{Yahoo!}):
\begin{center}
\begin{tabular}{cl}
    \texttt{"* that PATTERN *"} & (\emph{loose})\\
    \texttt{"HEAD that PATTERN *"} & (\emph{strict})\\
    \texttt{"* that PATTERN MOD"} & (\emph{strict})\\
\end{tabular}
\end{center}

\noindent where  \texttt{PATTERN} is  an inflected  form of  a verb,
\texttt{MOD}  and \texttt{HEAD} are inflected forms
the  modifier and the head of a noun compound that is paraphrasable by the pattern,
\texttt{that} is the  word \emph{that},
and  \texttt{*} is  the  search engine's  star operator.

We use the first  pattern for \emph{loose bootstrapping} and
the other  two for both \emph{strict  bootstrapping} and \emph{NC-only
strict bootstrapping}.

Note that the above queries  are generalizations of the actual queries
we  use against  the  search  engine. In  order  to instantiate  these
generalizations, we further generate  the possible inflections for the
verbs  and the  nouns involved.  For nouns,  we produce singular and
plural forms, while  for verbs, we vary not only  the number (singular
and plural), but  also the tense (we allow present,  past, and present
perfect). When  inflecting verbs,  we distinguish between  active verb
forms  like \emph{consist  of}  and passive  ones  like \emph{be  made
from} and  we treat  them accordingly.
Overall, in the case of  \emph{loose bootstrapping}, we generate about
14  and  20 queries  per  pattern  for  active and  passive  patterns,
respectively, while for  \emph{strict bootstrapping} and \emph{NC-only
strict  bootstrapping},  the  instantiations  yield about  28  and  40
queries for active and passive patterns, respectively.

For example, given the seed \emph{be made of},
we could generate
\texttt{"* that were made of *"}.
If we are further given the NC \emph{orange juice},
we could also produce
\texttt{"juice that was made of *"} and
\texttt{"* that is made of oranges"}.

\subsubsection{Snippet Extraction}
\label{sec:snippet:extract:step1}

We  execute  the  above-described instantiations  of  the  generalized
queries against a search engine as exact phrase queries, and, for each
one, we collect the snippets for the top 1,000 returned results.

\subsubsection{NC Extraction and Filtering}
\label{sec:NC:extract}

Next, we  process the snippets  returned by  the search engine  and we
acquire potential noun  compounds from them.
Then, in each  snippet, we look for  an instantiation
of the  pattern used  in the  query and  we try
to  extract  suitable  noun(s)  that occupy  the  position(s)  of  the
\texttt{*}.

For \emph{loose  bootstrapping}, we extract  two nouns, one  from each
end of the matched pattern,  while for \emph{strict bootstrapping} and
for  \emph{NC-only strict  bootstrapping}, we  only extract  one noun,
either preceding  or following  the pattern, since  the other  noun is
already fixed. We then lemmatize the  extracted noun(s) and
we  form  NC candidates  from  the  two arguments  of  the
instantiated  pattern,  taking into  account  whether  the pattern  is
active or passive.

Due  to  the   vast  number  of  snippets  we  have   to  process,  we
decided not to use a  syntactic parser  or a  part-of-speech (POS)
tagger\footnote{In  fact,  POS  taggers  and  parsers  are  unreliable
for  Web-derived snippets,  which often  represent parts  of sentences
and  contain errors  in  spelling,  capitalization and  punctuation.};
thus, we  use   heuristic  rules  instead.  We   extract  ``phrases''  using
simple  indicators   such  as   punctuation  (e.g.,   comma,  period),
coordinating conjunctions\footnote{Note  that filtering  the arguments
using  such  indicators  indirectly subsumes  the  pattern  \texttt{"X
PATTERN Y \underline{and}"}  proposed in \cite{Kozareva:2010}.} (e.g.,
\textit{and,  or}),   prepositions  (e.g.,  \textit{at,   of,  from}),
subordinating conjunctions (e.g.,  \textit{because, since, although}),
and  relative  pronouns (e.g.,  \textit{that,  which,  who}). We  then
extract the nouns from these phrases, we lemmatize them using WordNet,
and we form a list of NC candidates.

While the above heuristics work reasonably well in practice,
we perform some further filtering, removing all NC candidates
for which one or more of the following conditions are met:

\begin{enumerate}
\item the candidate NC is one of the seed examples or has been extracted on a previous iteration;
\item the head and the modifier are the same;
\item the head or the modifier are not both listed as nouns in WordNet \cite{Fellbaum:1998};
\item the candidate NC occurs less than 100 times in the \emph{Google Web 1T 5-gram corpus}\footnote{http://www.ldc.upenn.edu/Catalog/CatalogEntry.jsp?\\catalogId=LDC2006T13};
\item the NC is extracted less than $N$ times (we tried 5 and 10) in the context of the pattern for all instantiations of the pattern.
\end{enumerate}

\subsection{Bootstrapping Step 2: Pattern Extraction}
\label{sec:extraction-pattern}

This is  the second step  of our  bootstrapping algorithm as  shown on
\figref{fig:process}. Given a list of  noun compounds, we mine the Web
to  extract patterns:  verbs and/or  prepositions that  can paraphrase
each  NC.  The  idea is  to  turn the  NC's  pre-modifier into  a
post-modifying relative  clause and to  collect the verbs and prepositions
that  are used in such  clauses. Below we describe each of  the sub-steps of the
NC extraction  process: query  generation, snippet harvesting,  and NC
extraction \& filtering.

\subsubsection{Query Generation}

The  process of  extraction  starts with  exact-phrase queries  issued
against a Web search engine  (again \emph{Yahoo!}) using the following
generalized pattern:

\begin{center}
    \texttt{"HEAD THAT? * MOD"}
\end{center}

\noindent where \texttt{MOD} and  \texttt{HEAD} are inflected forms of
NC's  modifier  and  head,  respectively,  \texttt{THAT?}  stands  for
\emph{that},  \emph{which},  \emph{who}  or   the  empty  string,  and
\texttt{*} stands for 1-6 instances of search engine's star operator.

For example, given \emph{orange juice},
we could generate queries like
\texttt{"juice that * oranges"},
\texttt{"juices which * * * * * * oranges"},
and \texttt{"juices * * * orange"}.

\subsubsection{Snippet Extraction}

The same as in \secref{sec:snippet:extract:step1} above.

\subsubsection{Pattern Extraction and Filtering}
\label{sec:pattern:filtering}

We  split  the  extracted  snippets into  sentences,  and  filter  out
all  incomplete  ones  and  those  that do  not  contain  (a  possibly
inflected  version  of)  the  target   nouns.  We  further  make  sure
that  the   word  sequence  following  the   second  mentioned  target
noun  is   non-empty  and  contains   at  least  one   non-noun,  thus
ensuring  the  snippet  includes  the  entire  noun  phrase.  We  then
perform
shallow parsing,
and we extract all verb  forms, and the following preposition, between
the  target nouns.  We allow  for adjectives  and participles  to fall
between the verb and the preposition but not nouns; we further ignore
modal  verbs and  auxiliaries, but  we retain  the passive  \emph{be},
and  we  make sure  there  is  exactly  one  verb phrase  between  the
target nouns.  Finally, we  lemmatize the verbs  to form  the patterns
candidates, and we apply the following pattern selection rules:

\begin{enumerate}
\item we filter out all patterns that were provided as initial seeds or were extracted previously;
\item we select the top 20 most frequent patterns;
\item we filter out all patterns that were extracted less than $N$ times (we tried 5 and 10) and with less than $M$ NCs per pattern (we tried 20 and 50).
\end{enumerate}

\section{Target Relation and Seed Examples}
\label{sec:data}

\begin{table*}[htb]
\footnotesize
\begin{center}
\begin{tabular}{|l|}
\hline
\textbf{Seed NCs}: bronze statue, cable network, candy cigarette, chocolate bar, concrete desert, copper coin, daisy chain, glass eye, \\
immigrant minority, mountain range, paper money, plastic toy, sand dune, steel helmet, stone tool, student committee, \\
sugar cube, warrior castle, water drop, worker team \\
\hline
\textbf{Seed patterns}: be composed of, be comprised of, be inhabited by, be lived in by, be made from, be made of, be made up of, \\
be manufactured from, be printed on, consist of, contain, have, house, include, involve, look like, resemble, taste like\\
\hline
\end{tabular}
\caption{Our seed examples: 20 noun compounds and 18 verb patterns.}
\label{tab:seed}
\end{center}
\end{table*}

As  we mentioned  above, we  use the  inventory of  abstract relations
proposed   in   the   popular  theoretical   linguistics   theory   of
\newcite{Levi:1978}. In  this theory, noun compounds  are derived from
underlying relative clauses or noun phrase complement constructions by
means  of  two general  processes:  predicate  deletion and  predicate
nominalization.  Given   a  two-argument  predicate,   {\it  predicate
deletion} removes that predicate, but retains its arguments to form an
NC, e.g., {\it  pie made of apples} $\rightarrow$ {\it  apple pie}. In
contrast, {\it predicate nominalization} creates an NC whose head is a
nominalization  of  the underlying  predicate  and  whose modifier  is
either the  subject or the  object of  that predicate, e.g.,  {\it The
President  refused General  MacArthur's  request.} $\rightarrow$  {\it
presidential refusal}.

According  to Levi,  predicate  deletion can  be  applied to  abstract
predicates,  whose semantics  can be  roughly approximated  using five
paraphrasing  verbs   (\textsc{Cause},  \textsc{Have},  \textsc{Make},
\textsc{Use},  and \textsc{Be})  and  four prepositions  (\textsc{In},
\textsc{For},  \textsc{From},   and  \textsc{About}).

Typically,  in predicate deletion,  the modifier  is derived from  the object  of the
underlying relative clause; however, the  first three verbs also allow
for it to be derived from  the subject. Levi expresses the distinction
using  indexes. For  example,  \emph{music  box} is  \textsc{Make$_1$}
(object-derived), i.e., the  \emph{box \underline{makes} music}, while
\emph{chocolate  bar}  is \textsc{Make$_2$}  (subject-derived),  i.e.,
\emph{the bar \underline{is made of} chocolate} (note the passive).

Due  to  time constraints,  we  focused  on  one relation  of  Levi's,
\textsc{Make$_2$}, which  is among the  most frequent relations
an NC can express  and is present in some form in many relation inventories
\cite{warren:1978,Barker:1998,Rosario:2001,Nastase:Szpakowicz:2003,Girju:2005,Girju:2007:task4,girju09:lre,hendrickx-EtAl:2010:SemEval,Tratz:2010}.

In Levi's theory, \textsc{Make$_2$} means that the head of the noun compound is made up of
or is  a product of  its modifier. There  are three subtypes  of this
relation (we do not attempt to distinguish between them):

\begin{enumerate}
\item[(a)] the modifier is a unit and the head is a configuration, e.g., {\it root system};
\item[(b)] the modifier represents a material and the head is a mass or an artefact, e.g., {\it chocolate bar};
\item[(c)] the head represents human collectives and the modifier specifies their membership, e.g., {\it worker teams}.
\end{enumerate}

There  are  20  instances  of \textsc{Make$_2$}  in  the  appendix  of
\cite{Levi:1978},  and  we  use  them   all  as  \emph{seed  NCs}.  As
\emph{seed  patterns},   we  use   a  subset  of   the  human-proposed
paraphrasing verbs and  prepositions corresponding to these  20 NCs in
the dataset in \cite{Nakov:2008:AIMSA}, where each NC is paraphrased by
25-30 annotators. For  example, for \emph{chocolate bar},  we find the
following list  of verbs (the  number of annotators who  proposed each
verb is shown in parentheses):

\begin{quotation}
\begin{small}
\noindent be made of (16), contain (16), be made from (10), be composed of (7), taste like (7), consist of (5), be (3), have (2), melt into (2), be manufactured from (2), be formed from (2), smell of (2), be flavored with (1), sell (1), taste of (1), be constituted by (1), incorporate (1), serve (1), contain (1), store (1), be made with (1), be solidified from (1), be created from (1), be flavoured with (1), be comprised of (1).
\end{small}
\end{quotation}

As we can see, the most frequent patterns are of highest quality, e.g., \emph{be made of (16)},
while the less frequent ones can be wrong, e.g., \emph{serve (1)}.
Therefore, we filtered out all verbs that were proposed less than five
times with  the 20 seed  NCs. We  further removed the  verb \emph{be},
which is too general, thus ending  up with 18 seed patterns. Note that
some patterns  can paraphrase multiple  NCs: the total number  of seed
NC-pattern pairs is 84.

The seed NCs  and patterns are shown in  \tabref{tab:seed}. While some
patterns, e.g., \emph{taste  like} do not express  the target relation
\textsc{Make$_2$}, we  kept them  anyway since  they were  proposed by
several human  annotators and since  they do express  the fine-grained
semantics of some particular instances of that relation;
thus, we thought they might be useful, even for the general relation.
For  example,  \emph{taste  like}  has   been  proposed  8  times  for
\emph{candy cigarette}, 7 times for  \emph{chocolate bar}, and 2 times
for \emph{sugar cube},  and thus it clearly correlates  well with some
seed  examples,  even if  it  does  not express  \textsc{Make$_2$}  in
general.

\section{Experiments and Evaluation}
\label{sec:experiment}

Using  the  NCs and patterns in  \tabref{tab:seed}   as  initial  seeds,  we  ran
our  algorithm  for  three iterations  of  \emph{loose  bootstrapping}
and   \emph{strict  bootstrapping},   and   for   two  iterations   of
\emph{NC-only strict bootstrapping}. We only performed up to three iterations
because of the huge number of noun compounds extracted
for \emph{NC-only strict bootstrapping} (which we only ran for two iterations)
and because of the low number of new NCs extracted by \emph{loose bootstrapping} on iteration 3.
While we could have run \emph{strict bootstrapping} for more iterations,
we opted for a comparable number of iterations for all three methods.

\begin{table}[bth]
\begin{center}
\footnotesize
\begin{tabular}{|l|rrr|}
\hline
\textbf{Limits} & \multicolumn{3}{c|}{\bf Extracted \& Retained} \\
(see \ref{sec:pattern:filtering}) & \multicolumn{1}{c}{\bf NCs} & \multicolumn{1}{c}{\bf Patterns } & \multicolumn{1}{c|}{\bf Patt.+NC}\\
\hline
\multicolumn{4}{l}{\textbf{Loose Bootstrapping}}\\
\hline
$N$=5, $M$=50 & 1,662 / 61.67 & 12 / 65.83 & 1,337 \\
$N$=10, $M$=20 &   590 / 61.52 & 9 / 65.56 & 316 \\
\hline
\multicolumn{4}{l}{\textbf{Strict Bootstrapping}}\\
\hline
$N$=5, $M$=50 & 25,375 / 67.42 & 16 / 71.43 & 9,760 \\
$N$=10, $M$=20 & 16,090 / 68.27 & 16 / 78.98 & 5,026\\
\hline
\multicolumn{4}{l}{\textbf{NC-only Strict Bootstrapping}}\\
\hline
$N$=5  & 205,459 / 69.59 & -- & -- \\
$N$=10 & 100,550 / 70.43 & -- & -- \\
\hline
\end{tabular}
\caption{\label{tab:extracted:count}
Total number and accuracy in \%
for NCs, patterns and NC-pattern pairs extracted and retained
for each of the three methods over all iterations.
}
\end{center}
\end{table}

\begin{table*}[tbh]
\begin{center}
\footnotesize
\begin{tabular}{|l|c@{}c|@{}r@{ }@{ }r|r@{ }@{ }@{ }r|r@{ }@{ }@{ }r|}
\hline
\textbf{Limits} & \multicolumn{2}{c|}{\bf Seeds} & \multicolumn{2}{c|}{\bf Iteration 1} & \multicolumn{2}{c|}{\bf Iteration 2} & \multicolumn{2}{c|}{\bf Iteration 3} \\
(see \ref{sec:pattern:filtering}) & \multicolumn{1}{c}{Patt.} & \multicolumn{1}{c|}{NCs} & \multicolumn{1}{c}{Patt.} & \multicolumn{1}{c|}{NCs} & \multicolumn{1}{c}{Patterns} & \multicolumn{1}{c|}{NCs} & \multicolumn{1}{c}{Patterns} & \multicolumn{1}{c|}{NCs} \\
\hline
\multicolumn{9}{l}{\textbf{Loose Bootstrapping}}\\
\hline
$N$=5, $M$=50 & \multicolumn{1}{c}{--} & 18 & \multicolumn{1}{c}{--} & 1,144 / 63.11 & 1,136 / 64.44 / 9 & 390 / 58.72 & 201 / 70.00 / 3 & 128 / 57.03 \\
$N$=10, $M$=20 & \multicolumn{1}{c}{--} & 18 & \multicolumn{1}{c}{--} & 502 / 61.55 & 294 / 62.50 / 8 & 78 / 60.26 & 22 / 90.00 / 1 & 10 / 70.00 \\
\hline
\multicolumn{9}{l}{\textbf{Strict Bootstrapping}}\\
\hline
$N$=5, $M$=50  & 20 & 18 & \multicolumn{1}{c}{--} & 7,011 / 70.65 & 5,312 / 74.00 / 10 & 11,214 / 67.15 & 4,448 / 60.00 / 6 & 7,150 / 64.69 \\
$N$=10, $M$=20 & 20 & 18 & \multicolumn{1}{c}{--} & 4,826 / 71.26 & 2,838 / 79.38 / 10 & 7,371 / 67.26 & 2,188 / 78.33 / 6 & 3,893 / 66.48 \\
\hline
\multicolumn{9}{l}{\textbf{NC-only Strict Bootstrapping}}\\
\hline
$N$=5  & 20 & 18 & \multicolumn{1}{c}{--} & 7,011 / 70.65 & \multicolumn{1}{c}{--} & 198,448 / 69.55 & \multicolumn{1}{c}{--} & \multicolumn{1}{c|}{--} \\
$N$=10 & 20 & 18 & \multicolumn{1}{c}{--} & 4,826 / 71.26 & \multicolumn{1}{c}{--} &  95,524 / 70.59 & \multicolumn{1}{c}{--} & \multicolumn{1}{c|}{--} \\
\hline
\end{tabular}
\caption{\label{tab:results}
\textit{Evaluation results  for up to  three iterations.} For  NCs, we
show the number of unique NCs  extracted and their accuracy in \%. For
patterns, we  show the  number of  unique NC-pattern  pairs extracted,
their accuracy in  \%, and the number of unique  patterns retained and
used  to extract  NCs on  the second  step of  the current  iteration.
The  first column  shows the  pattern filtering  thresholds used  (see
\secref{sec:pattern:filtering} for details).
}
\end{center}
\end{table*}

Examples of noun compounds that we have extracted are \textit{bronze bell}  (be made of,  be made from)
and \textit{child team}  (be  composed  of,  include).
Example patterns  are \textit{be filled with} (cotton bag, water cup)
and \textit{use}  (water sculpture, wood  statue).

Tables    \ref{tab:extracted:count} and \ref{tab:results} show
the    overall     results.    As    we    mentioned     in    section
\ref{sec:pattern:filtering}, at  each iteration,  we filtered  out all
patterns that were extracted less than $N$ times or with less than $M$
NCs.
Note that
we only used the 10 most frequent NCs per pattern as NC seeds
for NC extraction in the next iteration
of \emph{strict bootstrapping} and \emph{NC-only strict bootstrapping}.
\tabref{tab:results} shows the results for two value combinations of ($N$;$M$): (5;50) and (10;20).
Note also that if some NC was extracted by  several different patterns, it was only counted once.
Patterns are subject to particular NCs,
and thus we show
(1)~the number of patterns extracted with all NCs, i.e., unique NC-pattern pairs,
(2)~the accuracy of these pairs,\footnote{One of the reviewers suggested that
evaluating the accuracy of NC-pattern pairs could potentially conceal some of the drift of our algorithm.
For example, while \emph{water cup} / \emph{be filled with} is a correct NC-pattern pair,
\emph{water cup} is incorrect for \textsc{Make$_2$};
it is probably an instance of Levi's \textsc{For}.
Thus, the same bootstrapping technique evaluated against a fixed set of semantic relations
(which is the more traditional approach)
could arguably show bootstrapping going ``off the rails'' more quickly
than what we observe here.
However, our goal, as stated in Section \ref{sec:representation},
is to find NC-specific paraphrases,
and our evaluation methodology is more adequate with respect to this goal.}
and
(3)~the number of unique patterns retained after filtering,
which will be used to extract new noun compounds on the second step of the current iteration.

The above accuracies were calculated based on  human judgments
by an experienced, well-trained annotator.
We also hired a second annotator for a small subset of the examples.

For NCs,  the first annotator  judged whether  each NC is  an instance
of  \textsc{Make$_2$}.  All  NCs  were judged,  except  for  iteration
2 of \emph{NC-only  strict bootstrapping},  where their  number was
prohibitively high and only the most frequent noun compounds extracted
for each modifier and for each  head were checked: 9,004 NCs for $N$=5
and 4,262 NCs for $N$=10.

For patterns, our first annotator judged the correctness of the unique
NC-pattern  pairs, i.e.,  whether the  NC is  paraphrasable with the
target pattern.  Given the large  number of NC-pattern pairs,
the  annotator only  judged patterns with  their top  10 most
frequent NCs.  For example, if  there were 5 patterns  extracted, then
the NC-pattern pairs to be judged would  be no more than 5 $\times$ 10
= 50.

Our  second annotator  judged  340  random examples:  100  NCs and  20
patterns with their  top 10 NCs for each iteration.  The Cohen's kappa \cite{cohen:1960:kappa}
between the  two annotators is  .66 (85\% initial  agreement), which
corresponds to substantial agreement \cite{Landis:1977:kappa}.

\section{Discussion}
\label{sec:discussion}

Tables  \ref{tab:extracted:count} and \ref{tab:results}  show  that
fixing one of the two nouns in the pattern, as  in
\emph{strict  bootstrapping} and  \emph{NC-only strict  bootstrapping},
yields significantly higher  accuracy ($\chi^2$ test) for  both NC and
NC-pattern pair extraction compared to \emph{loose bootstrapping}.

The  accuracy for \emph{NC-only strict bootstrapping}  is a
bit higher  than for \emph{strict bootstrapping},
but the actual differences are probably smaller
since the evaluation of the former on  iteration  2 was done for the  most frequent NCs, which are more accurate.

Note that the number of extracted NCs is much higher with the strict
methods because of the higher number of possible instantiations of the
generalized query  patterns. For \emph{NC-only  strict bootstrapping},
the number  of extracted NCs  grows exponentially since the  number of
patterns does not diminish as in the other two methods.
The number of extracted patterns  is similar for the different methods
since we select no more than 20 of them per iteration.

Overall, the accuracy for all  methods decreases from one iteration to
the next since errors 
accumulate;
still, the degradation is slow.
Note also the exception of \emph{loose bootstrapping} on iteration 3.

Comparing the results  for $N$=5 and $N$=10, we can  see that, for all
three  methods,  using  the  latter  yields  a  sizable  drop  in  the
number  of  extracted NCs  and  NC-pattern  pairs;  it also  tends  to
yield a  slightly improved accuracy.  Note, however, the  exception of
\emph{loose  bootstrapping} for  the first  two iterations,  where the
less restrictive $N$=5 is more accurate.

\begin{table}[htb]
\begin{center}
\footnotesize
\begin{tabular}{|l|r@{ }@{ }@{ }r@{ }@{ }@{ }r|r|}
\hline
\textbf{Rep.} & \textbf{Iter. 1} & \textbf{Iter. 2} & \textbf{Iter. 3} & \multicolumn{1}{c|}{\textbf{All}} \\
\hline
Syn. & 11/81.81  & 3/66.67    & 0        & 14/78.57 \\
Hyp. & 27/85.19  & 35/77.14   & 33/66.67 & 95/75.79 \\
Sis. & 381/82.05 & 1,736/69.33 & 17/52.94 & 2,134/75.12 \\
\hline
All  & 419/82.58 & 1,774/71.68 & 50/62.00 & 2,243/\textbf{75.47} \\
\hline
\end{tabular}
\caption{Number of extracted noun compounds and accuracy in \% for the method of \newcite{Kim:2007}.
The abbreviations \textit{Syn.}, \textit{Hyp.}, and \textit{Sis.}
indicate using synonyms, hypernyms, and sister words, respectively.}
\label{tab:kim07}
\end{center}
\end{table}

As  a comparison,  we  implemented the  method of  \newcite{Kim:2007},
which generates  new semantically interpreted NCs  by replacing either
the  head  or the  modifier  of  a  seed  NC with  suitable  synonyms,
hypernyms  and  sister  words  from WordNet,
followed by similarity filtering using \texttt{WordNet::Similarity}
\cite{Pedersen:2004:wordnetsim}.

The  results  for  three
bootstrapping iterations using the same list of 20 initial seed NCs as
in our previous  experiments, are shown in  \tabref{tab:kim07}. We can
see that the overall accuracy of  their method is slightly better than
ours. Note, however, that our method  acquired a much larger number of
NCs, while  allowing more variety  in the NC semantics.  Moreover, for
each extracted noun compound, we also generated a list of fine-grained
paraphrasing verbs.

\section{Error Analysis}
\label{sec:erroranalysis}

Below we analyze the errors of our method.

Many problems were  due to wrong POS assignment. For  example, on Step
2, because  of the omission  of \textit{that} in  ``\textit{the statue
has such  high quality gold  (that) demand is  ...}'', \textit{demand}
was tagged as a  noun and thus extracted as an  NC modifier instead of
\textit{gold}. The problem also arose on Step 1, where we used WordNet
to check whether  the NC candidates were composed of  two nouns. Since
words like \textit{clear},  \textit{friendly}, and \textit{single} are
listed in  WordNet as nouns (which  is possible in some  contexts), we
extracted  wrong NCs  such  as  \textit{clear cube},  \textit{friendly
team},  and  \textit{single chain}.  There  were  similar issues  with
verb-particle constructions since some particles  can be used as nouns
as well, e.g., \textit{give \underline{back}, break \underline{down}}.

Some errors were due to semantic transparency issues,
where the syntactic and the semantic head of a target NP were mismatched
\cite{Fillmore:2002:Seeing:arguments,Fontenelle:1999}.
For  example, from  the  sentence  ``\emph{This wine  is  made from  a
\underline{range of  white grapes}.}'', we would  extract \emph{range}
rather than \emph{grapes} as the potential modifier of \emph{wine}.

In some  cases, the NC-pattern  pair was correct,  but the NC  did not
express the  target relation, e.g.,  while \textit{contain} is  a good
paraphrase for  \textit{toy box}, the  noun compound itself is  not an
instance of \textsc{Make$_2$}.

\begin{figure}[htb]
\begin{center}
\includegraphics[width=2.6in]{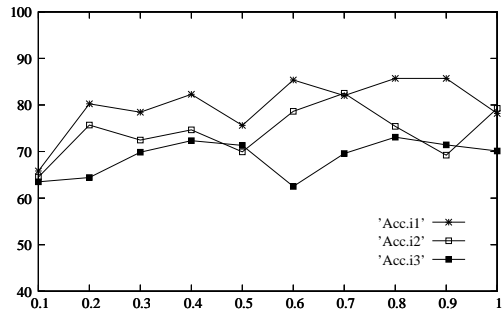}
\end{center}
\caption{NC accuracy vs.\ collocation strength.
\label{fig:compositionality2}
}
\end{figure}

There were also cases where the pair of extracted nouns did not make a
good NC,  e.g., \textit{worker work}  or \textit{year toy}.  Note that
this is despite  our checking that the candidate NC  occurred at least
100  times in  the \emph{Google  Web  1T 5-gram  corpus} (see  Section
\ref{sec:NC:extract}). We hypothesized that such bad NCs would tend to
have  a low  collocation  strength. We  tested  this hypothesis  using
the  Dice  coefficient,  calculated  using  the  \emph{Google  Web  1T
5-gram corpus}. \figref{fig:compositionality2} shows  a plot of the NC
accuracy vs. collocation strength for \emph{strict bootstrapping} with
$N$=5,  $M$=50 for  all three  iterations (the  results for  the other
experiments  show a  similar  trend).  We can  see  that the  accuracy
improves slightly as the collocation strength increases:
compare the left and the right ends of the graph
(the results are mixed in the middle though).

\section{Conclusion and Future Work}
\label{sec:conclusion}

We have  presented a framework  for building  a very large  dataset of
noun compounds  expressing a given target  abstract semantic relation.
For  each  extracted  noun  compound,  we  generated  a  corresponding
fine-grained  semantic interpretation:  a frequency  distribution over
suitable paraphrasing verbs.

In  future work,  we  plan to  apply our  framework  to the  remaining
relations in  the inventory of \newcite{Levi:1978},  and to release
the  resulting dataset  to  the research  community.  We believe  that
having a large-scale  dataset of noun compounds  interpreted with both
fine-  and coarse-grained  semantic  relations would  be an  important
contribution to  the debate  about which representation  is preferable
for different tasks. It should also help the overall advancement of the
field of noun compound interpretation.


\section*{Acknowledgments}

This research is partially supported (for the second author) by the \emph{SmartBook project}, funded by the Bulgarian National Science Fund under Grant D002-111/15.12.2008.

We would like to thank the anonymous
reviewers for their detailed and constructive
comments, which have helped us improve the paper.

\bibliographystyle{acl}
\bibliography{ncpattern}

\end{document}